\documentclass{ieeeaccess}
\usepackage{cite}
\usepackage{amsmath,amssymb,amsfonts}
\usepackage{algorithmic}
\usepackage{graphicx}
\usepackage{textcomp}
\usepackage{url}

\usepackage{bm}
\makeatletter
\AtBeginDocument{\DeclareMathVersion{bold}
\SetSymbolFont{operators}{bold}{T1}{times}{b}{n}
\SetSymbolFont{NewLetters}{bold}{T1}{times}{b}{it}
\SetMathAlphabet{\mathrm}{bold}{T1}{times}{b}{n}
\SetMathAlphabet{\mathit}{bold}{T1}{times}{b}{it}
\SetMathAlphabet{\mathbf}{bold}{T1}{times}{b}{n}
\SetMathAlphabet{\mathtt}{bold}{OT1}{pcr}{b}{n}
\SetSymbolFont{symbols}{bold}{OMS}{cmsy}{b}{n}
\renewcommand\boldmath{\@nomath\boldmath\mathversion{bold}}}
\makeatother

\def\BibTeX{{\rm B\kern-.05em{\sc i\kern-.025em b}\kern-.08em
    T\kern-.1667em\lower.7ex\hbox{E}\kern-.125emX}}

\begin{document}
\history{Date of publication Feb 02, 2026, date of current version Feb 02, 2026.}
\doi{10.1109/ACCESS.2017.DOI}

\title{Large Language Models in the Abuse Detection Pipeline}
\author{\uppercase{Suraj Kath}\authorrefmark{1}, 
\uppercase{Sanket Badhe}\authorrefmark{1},
\uppercase{Preet Shah}\authorrefmark{1},
\uppercase{Ashwin Sampathkumar}\authorrefmark{1},
\uppercase{Shivani Gupta}\authorrefmark{1}}

\address[1]{Google LLC}

\markboth
{Kath \textit{et al.}: Large Language Models in the Abuse Detection Pipeline}
{Kath \textit{et al.}: Large Language Models in the Abuse Detection Pipeline}

\corresp{Corresponding author: Suraj Kath (e-mail: surajka@google.com).}

\begin{abstract}
Online abuse has grown increasingly complex, spanning toxic language, harassment, manipulation, and fraudulent behavior. Traditional machine-learning approaches dependent on static classifiers and labor-intensive labeling struggle to keep pace with evolving threat patterns and nuanced policy requirements. Large Language Models introduce new capabilities for contextual reasoning, policy interpretation, explanation generation, and cross-modal understanding, enabling them to support multiple stages of modern safety systems. This survey provides a lifecycle-oriented analysis of how LLMs are being integrated into the Abuse Detection Lifecycle (ADL), which we define across four stages: (I) Label \& Feature Generation, (II) Detection, (III) Review \& Appeals, and (IV) Auditing \& Governance. For each stage, we synthesize emerging research and industry practices, highlight architectural considerations for production deployment, and examine the strengths and limitations of LLM-driven approaches. We conclude by outlining key challenges including latency, cost-efficiency, determinism, adversarial robustness, and fairness and discuss future research directions needed to operationalize LLMs as reliable, accountable components of large-scale abuse-detection and governance systems.
\end{abstract}


\titlepgskip=-15pt

\maketitle

\section{Introduction}
\label{sec:introduction}
\subsection{The Changing Face of Online Abuse}
Digital platforms today are fighting a constant and evolving battle against online abuse. In the past, moderation was often about finding bad words i.e. checking for specific slurs or insults. Today, however, the landscape is far more complex. Harmful behavior has expanded to include sophisticated harassment, financial scams, impersonation, and coordinated manipulation \cite{Zhang2025GuardiansAO}.

These threats evolve quickly, often changing faster than traditional safety systems can adapt. For years, the industry relied on static machine learning models, such as BERT-based classifiers \cite{caselli-etal-2021-hatebert}. These systems depend heavily on humans to label thousands of examples by hand, which is slow and resource-intensive \cite{10.1145/3604237.3626891}. Furthermore, these older models often struggle with nuance; they might catch a specific insult but miss sarcasm, cultural context, or dog whistles that break the rules without using obvious toxic words \cite{10976181}. As platforms grow globally, these limitations make it increasingly difficult to keep users safe from implicit harm \cite{elsherief-etal-2021-latent}.

\subsection{The LLM Paradigm Shift}
Large Language Models (LLMs) have introduced a fundamental shift in how we approach this problem. Unlike traditional models that simply look for patterns, LLMs can read a platform’s safety policy, reason about the intent behind a post, and even explain why it breaks the rules \cite{CHEN2025114689}. However, this technological leap creates a paradox: the same tools that help defenders are also arming attackers. The widespread availability of LLMs has lowered the barrier for malicious actors to generate unique, scalable, and personalized abuse, effectively raising the bar for what defense systems must catch \cite{Zhang2025GuardiansAO, BARMAN2024100545}.

To meet this challenge, LLMs are being integrated not just as detectors, but throughout the entire Abuse Detection Lifecycle (ADL) from creating training data to helping human reviewers make fair decisions \cite{CHEN2025114689}. This moves us away from rigid, single-model systems toward adaptive pipelines that are more policy-aware and accountable \cite{huang2025content}.

\subsection{Scope and Contributions}
While many recent surveys focus on benchmarking LLMs for specific tasks, such as toxicity classification, this paper takes a system-level view. We structure our analysis around the Abuse Detection Lifecycle (ADL), a framework that categorizes how LLMs are used across the full operational pipeline of Trust \& Safety.

Our analysis begins with Label and Feature Generation, exploring how LLMs create synthetic data and assist in labeling to reduce reliance on slow manual work \cite{horych-etal-2025-promises, 10.1145/3604237.3626891}. We then examine Detection, discussing how to use LLMs in real-time systems to catch abuse without slowing down the platform, often by combining them with faster, smaller models \cite{inan2023llamaguardllmbasedinputoutput}. The survey continues into the Review and Appeals stage, detailing how LLMs support human moderators by summarizing evidence and writing clear explanations for decisions \cite{di2025detection}. Finally, we analyze Auditing and Governance, where LLMs function as auditors to constantly check the system for fairness, bias, and consistency \cite{binh2025transforming, ijcai2024p801}.

In doing so, this paper makes several key contributions. We propose a unified framework to characterize LLM applications across the pipeline and provide a synthesis of real-world engineering trade-offs, such as cost, latency, and reliability \cite{10.1145/3604237.3626891}. Furthermore, we outline critical open challenges, including model fairness, adversarial robustness against jailbreaks \cite{10835584}, and the need for new governance standards to operationalize these models responsibly.

\subsection{Organization of this Survey}
The remainder of this paper is organized to follow the lifecycle of a safety system. Section 2 defines the Abuse Detection Lifecycle (ADL) framework. Section 3 explores the Cold Start problem and how LLMs assist in data labeling and generation. Section 4 examines detection architectures, moving from zero-shot classifiers to specialized fine-tuned models. Section 5 discusses the post-detection phase, focusing on explainability and appeals. Section 6 reviews auditing methodologies for fairness and robustness. Finally, Section 7 discusses the engineering challenges of production deployment and future directions.

\section{The Abuse Detection Lifecycle (ADL) Framework}

To fully understand the impact of LLMs on platform safety, it is necessary to move beyond viewing them as isolated detection tools. In real-world operations, abuse detection is not a single decision point but a continuous, iterative process involving data creation, real-time analysis, human review, and system oversight. We introduce the Abuse Detection Lifecycle (ADL) as a framework to organize these distinct roles, allowing us to characterize how LLMs function not just as classifiers, but as versatile assistants across the entire safety pipeline.

We define the ADL across four interconnected stages: Label and Feature Generation, Detection and Triage, Review and Appeals, and Auditing and Governance.

\begin{figure*}[t]
    \centering
    \includegraphics[width=\textwidth, height=0.75\textheight, keepaspectratio]{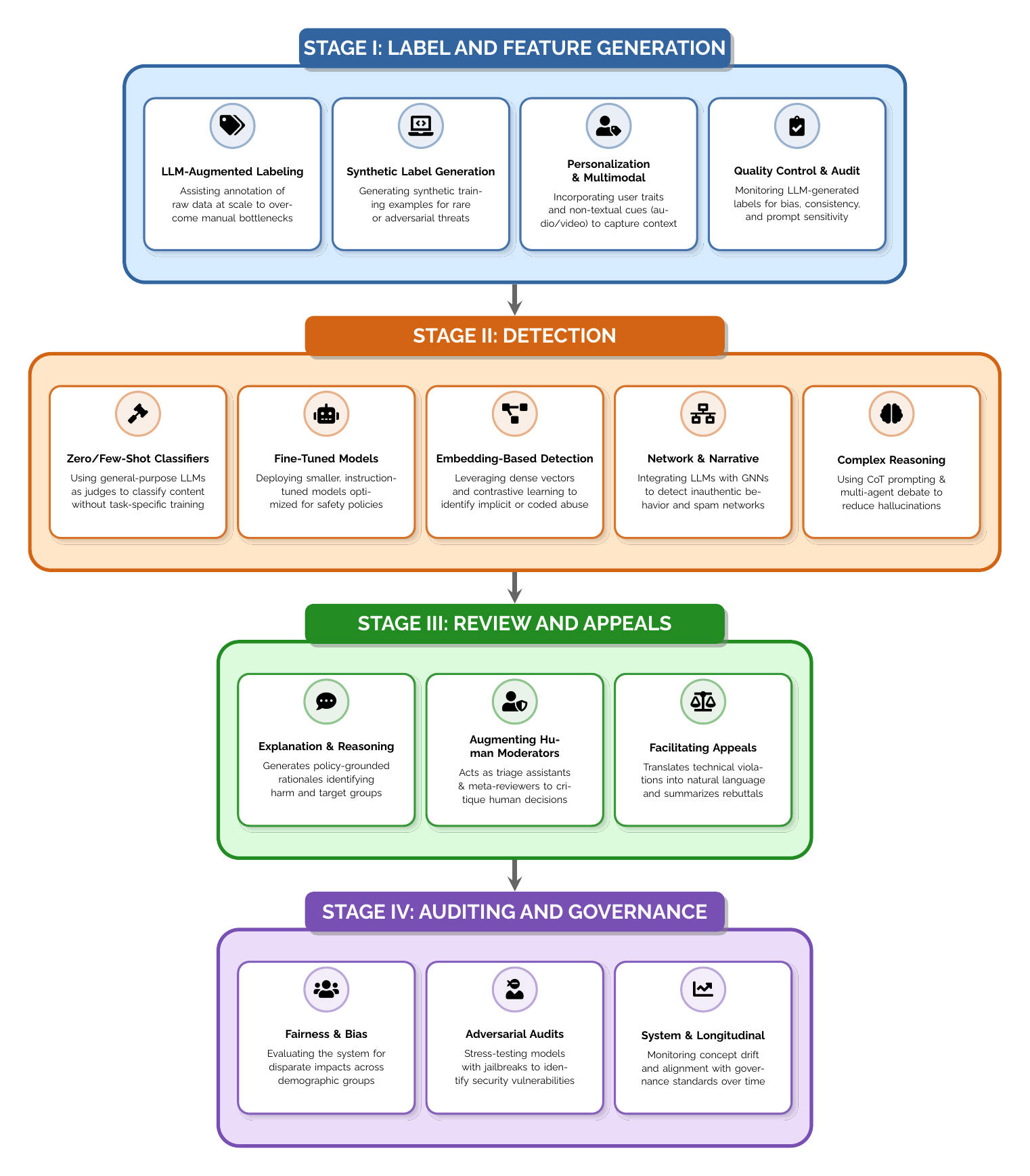}
    \caption{The Abuse Detection Lifecycle (ADL) Framework}
    \label{fig:adl_framework}
\end{figure*}

The first stage, Label and Feature Generation, addresses the Cold Start problem. Before any system can detect abuse, it requires labeled examples to learn from. Historically, this has been a major bottleneck, as human moderators cannot generate training data fast enough to keep up with rapidly evolving threats like new slang or coded hate speech. In the ADL framework, LLMs transform this stage by acting as Data Amplifiers. They assist by generating synthetic examples of emerging threats or by labeling raw data at scale, allowing safety teams to update their defenses rapidly without waiting for weeks of manual annotation.

Once the system is trained, it moves to Detection, the active defense stage where live content is analyzed. The central challenge here is volume; checking every user post with a massive, reasoning-heavy LLM is often too slow and expensive for large platforms. Consequently, LLMs in this stage typically function as Specialized Experts rather than general filters. They are deployed selectively to handle the grey area cases such as sarcasm, cultural nuances, or complex harassment that smaller, faster models fail to understand, ensuring that nuanced abuse is caught without slowing down the entire platform.

When content is flagged, it proceeds to the Review and Appeals stage, which focuses on the human element of moderation. Human reviewers are often overwhelmed by the sheer volume of cases and the difficulty of interpreting complex policy guidelines consistently. Here, LLMs serve as Cognitive Support Assistants. Rather than simply replacing human judgment, they support it by summarizing lengthy evidence, such as video transcripts or long comment threads, and by drafting clear, policy-grounded explanations for why a specific decision was made. This improves the consistency of enforcement and provides users with better transparency during the appeals process.

Finally, the lifecycle closes with Auditing and Governance, the oversight stage that ensures the system remains fair and effective over time. Safety systems can silently become biased or outdated as language and social norms evolve. In this final stage, LLMs act as Auditors and Meta-Evaluators. They are used to continuously stress test the system with adversarial attacks to find weaknesses and to analyze historical decisions for potential bias against specific demographic groups. This continuous feedback loop ensures that the abuse detection pipeline remains robust, accountable, and aligned with the platform's safety goals.

\section{Stage I: Label and Feature Generation (The Cold Start Problem)}

LLMs are utilized to create and augment high-quality data, particularly where human labeling is difficult, slow, or expensive.

\subsection{Importance and Challenges of Labeling}
Labeling abusive content remains inherently difficult due to ambiguity, subjectivity, and the critical role of context. Irfan (2025) shows that utterances such as ``Sorry, I cannot do it for you'' can be perceived as neutral or abusive depending entirely on the emotional tone in which they are delivered, demonstrating that purely textual labels often fail to capture communicative intent \cite{10976181}. This multimodal dependency illustrates why text-only annotation is unreliable and why abusive interactions are intrinsically ambiguous.

Hartvigsen (2022) provides complementary evidence: because the dataset consists predominantly of implicitly toxic statements, human annotators show only moderate agreement on labeling harmfulness, intent, or stereotyping \cite{hartvigsen-etal-2022-toxigen}. This demonstrates that annotators struggle most when abuse is expressed subtly rather than explicitly, reinforcing the challenge of producing reliable labels for real-world abusive language.

Subjectivity also arises from individual differences among annotators. Yao (2024) demonstrates that perceived harm varies systematically across psychological traits: annotators with strong Self-Down tendencies label significantly more messages as harmful, whereas more rational annotators tend to rate the same messages as neutral \cite{yao-etal-2024-personalised} . These structural differences reveal why labeling abusive content, especially implicit or context-dependent harms, remains one of the most challenging tasks in NLP.

\subsection{From Manual to LLM-Augmented Labeling}
LLMs increasingly augment manual labeling through scalable label generation. Horych (2025) use multiple LLMs with majority-vote aggregation to produce 48,330 synthetic media-bias labels, an order of magnitude larger than existing human-labeled datasets such as BASIL and BABE \cite{horych-etal-2025-promises}. This aggregation-of-subjectivity approach is directly transferable to abuse detection, where individual annotator bias is a primary bottleneck. Classifiers trained solely on these LLM-generated labels achieve performance comparable to human-annotated baselines, specifically performing close to or outperforming models trained on the expert-labeled BABE and BASIL datasets while simultaneously surpassing the zero-shot performance of their own 'teacher' LLMs by 5–9\% in Matthews Correlation Coefficient (MCC).

Cost-efficiency is equally critical for monitoring high-volume abusive streams. In the financial text-classification domain, Loukas (2024) further demonstrate that curated few-shot examples and retrieval-augmented prompting dramatically improve LLM annotation cost-efficiency \cite{10.1145/3604237.3626891}. Their RAG approach retrieves only 2.2\% of examples while maintaining or surpassing GPT-4’s few-shot accuracy resulting in significant cost savings (e.g., around \$700 for the Banking77 test set) highlighting that prompt design and example selection are central to cost-effective LLM labeling. For abuse detection, this suggests that RAG can enable continuous, low-cost monitoring of toxicity without the prohibitive expense of querying frontier models for every comment.

Rocha (2025) extend this paradigm by demonstrating that LLMs can generate highly structured annotations specifically QUESTION–ANSWER–TEXT REGION tuples for document understanding tasks \cite{rocha2025data}. Their instruction-guided annotation pipeline significantly reduces manual workload by leveraging detailed positive/negative examples and tailored generation rules.

\subsection{Synthetic Label Generation and Weak Supervision}
Synthetic labeling has become a practical solution when abusive datasets are scarce, costly, or difficult to annotate reliably. Horych (2025) show that ensemble-based synthetic labeling can scale far beyond typical manually annotated datasets \cite{horych-etal-2025-promises}. Using three LLMs as independent annotators and aggregating their labels through majority voting, they produce 48,330 synthetic media-bias labels. Classifiers trained on these synthetic labels not only match human-annotated baselines but also surpass their individual LLM teacher annotators on several downstream metrics, illustrating the promise of large-scale weak supervision via LLM consensus.

Hartvigsen (2022) is a leading example of machine-generated supervision for toxicity detection \cite{hartvigsen-etal-2022-toxigen}. Using GPT-3 and an adversarial generator (ALICE), the authors construct a dataset of 274,000 synthetic toxic and benign statements balanced across demographic groups. Models fine-tuned on this synthetic corpus achieve substantial performance improvements; particularly on implicit toxicity that human annotators struggle to identify. This work shows that synthetic data, when properly designed and filtered, can serve as high-quality weak supervision for downstream classifiers.

Meguellati (2025) introduce a complementary form of weak supervision through semantic augmentation, where LLMs enrich each training instance with context-bearing explanations, cleaned captions, or lists of harmful triggers \cite{Meguellati_Zeghina_Sadiq_Demartini_2025}. Rather than generating new synthetic samples, their method produces structured auxiliary signals that function as pseudo-labels, guiding downstream models toward the correct interpretation of subtle persuasive or toxic cues. Across persuasive memes, toxic comments, and hateful memes, these LLM-generated semantic annotations yield performance comparable to human-written explanations, demonstrating that augmenting existing samples with LLM-produced contextual cues can be as effective as classical synthetic data generation while being dramatically more cost-efficient.

Cost-efficient synthetic labeling is further explored in \cite{10.1145/3604237.3626891}, which examines LLM-generated examples for financial text classification. While GPT-4-generated synthetic instances can improve model performance, the paper observed diminishing marginal gains as more synthetic examples are added, highlighting the importance of quality over quantity. The retrieval-augmented prompting strategy reduces the dependence on large seed datasets by selecting only the most semantically relevant examples, substantially lowering inference costs while preserving accuracy.

In multimodal document settings, Rocha (2025) demonstrate that LLMs can generate structured QUESTION–ANSWER–TEXT REGION tuples entirely synthetically through a multi-stage pipeline that combines OCR, document layout analysis, controlled QA generation, and LLM-based filtering \cite{rocha2025data}. Although the paper does not evaluate downstream model training, it provides a strong proof-of-concept that high-structure synthetic annotations can be produced reliably when rigorous filtering is applied.

Finally, Yulia Kumar (2024) generated synthetic biased and cyberbullying text across demographic categories using leading LLMs such as ChatGPT-4o, Claude 3 Opus, Gemini-1.5, and Pi AI \cite{202407.0411}. Their results show that while LLMs can produce diverse synthetic abusive content, the generated data often reflect each model’s safety constraints and latent biases, including reluctance to generate certain categories of harm and skewed sentiment distributions. These findings underscore that synthetic abuse-related data can inadvertently encode or amplify preexisting LLM biases, making careful validation essential before use in weak supervision.

\subsection{Human-in-the-Loop Annotation and Quality Control}
Despite significant advances in automated labeling, human oversight continues to play a central role in ensuring the reliability and trustworthiness of abusive-content datasets. In practice, automation often shifts rather than eliminates the need for human judgment.

While Rocha (2025) proposes a fully automated annotation pipeline for document QA using LLMs both to generate and to evaluate QUESTION–ANSWER–TEXT REGION tuples, the authors acknowledge that strong models can still hallucinate details or misinterpret document structure \cite{rocha2025data}. Their work shows how automated LLM-based judgment functions as a proxy for human review, reducing but not entirely eliminating the need for targeted manual inspection when errors propagate.

Hartvigsen (2022) illustrates this dynamic clearly: although most content is machine-generated, human evaluators were essential to validate toxicity labels, intent, stereotyping, and group references \cite{hartvigsen-etal-2022-toxigen}. Reviewers corrected misclassifications from GPT-3 and verified adversarial examples produced by ALICE, showing that automated systems alone cannot guarantee dataset quality particularly for subtle or borderline cases.

In contrast, Yulia Kumar(2024) directly demonstrates why human quality control remains essential when generating synthetic abusive-language datasets \cite{202407.0411}. Leading LLMs vary widely in their safety behaviors: some refuse to produce harmful content, while others readily generate highly toxic or biased examples when prompted or jailbroken. These inconsistencies create skewed or incomplete synthetic corpora, requiring human review to remove inappropriate outputs and correct demographic or sentiment bias introduced by the generator model.

A complementary perspective comes from LLM-Sentry\cite{10835584}. Here, human reviewers iteratively expand and refine a curated HarmfulKB, which is a knowledge base of malicious and benign prompts used to detect jailbreak attempts. This incremental update loop enables rapid incorporation of new adversarial strategies without retraining. Although designed for model safety rather than dataset construction, its architecture characterized by continuous human amendments, escalation pathways for ambiguous cases, and iterative updates, closely mirrors best practices in annotation quality-control pipelines.

LLM-based semantic augmentation also exposes new quality-control considerations. Meguellati (2025) shows that LLMs frequently sanitize or censor toxic expressions when generating explanations, omitting precisely the harmful cues needed for training \cite{Meguellati_Zeghina_Sadiq_Demartini_2025}. Their study resolves this through carefully engineered prompts and by comparing LLM outputs against human-written explanations, effectively using human annotations as a calibration signal. This aligns with the broader lesson that, even in highly automated pipelines, human oversight remains essential for verifying that LLM-generated annotations preserve task-critical information rather than filtering it away.

Collectively, these works highlight a consistent insight: LLMs can automate large portions of labeling, but human-in-the-loop mechanisms remain indispensable for catching model failures, mitigating bias, and maintaining high-quality datasets for abuse detection.

\subsection{Personalization and Contextual Labeling}
Perceptions of abusive language vary significantly across individuals, shaped by psychological traits, prior experiences, and subjective sensitivity. Yao (2024) shows that these individual differences can be incorporated directly into LLM-based labeling systems \cite{yao-etal-2024-personalised}. The authors condition GPT-3.5 Turbo on user-specific attributes such as Rationality, Irrationality, and Self-Down etc. derived from psychological scales, and supply these traits to the model through a retrieval-augmented prompting framework enriched with mined decision rules. This personalized approach yields substantial improvements, producing 1.5–4.4\% higher weighted F1 scores and identifying more harmful messages for vulnerable user profiles.

Their findings illustrate that personalized contextual labeling allows LLMs to capture subtle differences in perceived harm. A single message may be judged as safe by one user profile and harmful by another, a divergence that the baseline model entirely misses. However, the experiments also reveal limitations: personalization can sometimes destabilize predictions when message content and user traits strongly contradict each other. For example, profane messages paired with attributes associated with neutral interpretations caused the system to flip classifications in up to 9\% of cases. While the authors do not frame these issues as fairness risks, the results imply that over-personalization may introduce unintended biases, leading to inconsistent or overly sensitive detection for certain groups. These observations underscore the need for safeguards and calibration to ensure equitable and consistent personalized labeling across diverse populations.

\subsection{Multimodal and Real-World Labeling}
Real abusive communication often involves not just textual content, but also tone, pitch, and conversational flow. Irfan (2025) proposes a multimodal framework that combines audio-based emotion signals with textual semantics: MFCC and Mel-spectrogram features are used for emotion detection, while the transcribed speech is encoded using SBERT embeddings \cite{10976181}. On a custom dataset of short, labeled conversations (Red Flag vs Green Flag), their proposed model achieves 92\% accuracy, outperforming text-only BERT and SBERT baselines (86\% and 88\% respectively), representing a gain of roughly 4–6 accuracy points that highlights the value of incorporating vocal cues alongside text. Although the paper does not isolate performance specifically on tone-dependent examples, its design is motivated by precisely those cases where spoken delivery conveys hostility or abuse that is not apparent from the text alone. More broadly, such multimodal strategies are crucial for real-world abusive interactions, where intent and perceived harm are often communicated through voice and paralinguistic cues rather than explicit textual content alone.

\subsection{Label Quality, Bias, and Auditability}
LLM-generated labels are susceptible to systematic biases that arise from model training and alignment. Zhang (2025) show that instruction-tuned models such as Flan-T5 and OPT-IML tend to under-predict abusive labels, which the authors link to label-imbalanced, negative-majority training corpora \cite{zhang2025llm}. In contrast, RLHF-aligned models such as LLaMA-2-Chat and GPT-3.5 exhibit the opposite tendency, over-predicting abusive labels, which the authors attribute to safety-oriented RLHF tuning that encourages models to err on the side of caution. These prediction biases are consistent in direction across multiple datasets and prompt configurations, underscoring the need for explicit calibration and bias audits in LLM-based annotation.

Horych (2025) further reveal ideological skew in synthetic media-bias labels, with different LLM annotators exhibiting distinct political leanings \cite{horych-etal-2025-promises}. Yao (2024) shows that personalization can cause the same abusive message to be labeled differently across user profiles, and in some cases messages labeled as harmful in the dataset are incorrectly classified as neutral when paired with certain user attributes \cite{yao-etal-2024-personalised}. Yulia Kumar (2024) expose additional forms of bias in synthetic cyberbullying corpora: some models generate overly aggressive or stereotyped content, while others refuse to produce harmful text altogether, leading to both toxicity amplification and censorship-like effects \cite{202407.0411}.

Hartvigsen (2022) is a central example in this area: toxicity classifiers commonly over-detect benign statements mentioning minority groups and under-detect implicitly harmful content \cite{hartvigsen-etal-2022-toxigen}. The dataset exposes spurious correlations, such as associating demographic terms with toxicity, that many models rely on inadvertently. This makes TOXIGEN especially useful for auditing demographic and representational biases in toxicity detection systems.

Jaremko (2025) contributes additional insights into the auditability of LLM-generated labels \cite{jaremko-etal-2025-revisiting}. The study shows that model predictions for implicit abuse vary widely across prompt styles, dataset types, and supervision settings (zero-shot vs. few-shot). The same model can rate an implicitly abusive sentence as harmful under one prompt template and neutral under another. The authors also find that LLMs often rely on surface cues, leading to both false positives (e.g., flagging benign statements mentioning sensitive groups) and false negatives (e.g., missing euphemistic or indirect harm). These findings highlight the need to evaluate LLM-based labels across multiple prompt configurations and datasets to detect prompt-sensitivity, spurious correlations, and overfitting to annotation artifacts.

Di Bonaventura (2025) shows that even instruction-tuned LLMs exhibit systematic biases in both classification and explanation quality \cite{di-bonaventura-etal-2025-detection}. Across zero-shot, few-shot, and knowledge-guided learning strategies, models frequently mishandle stereotypes, overreact to sensitive terms, and generate explanations containing vagueness, cultural assumptions, or reasoning errors. Their knowledge-guided variant, GLlama Alarm, reduces these failures and improves alignment with expert judgments, highlighting that explanation auditing is essential, since flawed rationales can quietly propagate bias into downstream abusive-language detectors.

Taken together, these studies highlight that LLM-based annotation pipelines require systematic auditing including category-wise error analysis, false-negative and false-positive profiling, and, where applicable, demographic and ideological calibration to ensure that generated labels are both high quality and trustworthy for downstream abuse detection.

\subsection{Toward Adaptive and Ethical Labeling Pipelines}
Scaling LLM-based annotation requires pipelines that are adaptive, safety-aware, and continuously monitored. Irtiza (2024) demonstrates how human-curated knowledge bases, escalation pathways, and iterative updates help systems stay aligned as new adversarial behaviors emerge, offering an alternative to static annotation guidelines that quickly become outdated \cite{10835584}. Rocha (2025) shows that structured annotation pipelines, built around detailed instructions, explicit positive and negative examples, and multi-stage validation, can reduce hallucinations and promote more consistent outputs across models \cite{rocha2025data}. Although the paper does not explicitly frame these mechanisms in ethical terms, its design principles illustrate how careful prompt structuring and layered checking are essential for reliable automated annotation.

Yulia Kumar (2024) highlights several ethical concerns in synthetic harmful-content generation, including toxicity amplification, biased sampling across demographic groups, and model refusal due to safety filters \cite{202407.0411}. These findings point to the need for safety-aware constraints, post-generation audits, and potentially multi-model consensus to prevent skewed or unsafe synthetic datasets.

Hartvigsen (2022) dataset also illustrates both the promise and the ethical challenges of synthetic generation \cite{hartvigsen-etal-2022-toxigen}. While its large-scale, implicitly toxic examples enable improved classifier performance, the study notes clear risks: machine-generated statements about minority groups can inadvertently reinforce stereotypes or be misused outside controlled research environments. The authors discuss policy implications and emphasize controlled deployment, making TOXIGEN a key reference point for understanding ethical considerations in synthetic harmful-content generation.

Finally, works such as \cite{yao-etal-2024-personalised} and \cite{zhang2025llm} show that annotation pipelines must balance personalization, fairness, and alignment bias. Both papers reveal circumstances in which labeling strategies may systematically over-detect or under-detect harm for certain demographic or psychological subgroups, underscoring the need for calibration and ongoing bias monitoring in LLM-driven labeling workflows.

\section{Stage II: Detection}

The LLM era marked a definitive transition in Anti Abuse detection architecture. Prior to this era, abuse detection relied heavily on supervised learning models such as BERT or RoBERTa, which were fine-tuned on static datasets of hate speech and toxicity \cite{caselli-etal-2021-hatebert}. While effective at identifying explicit slurs, these models struggled with implicit abuse such as sarcasm, dog-whistles, and context-dependent harassment and lacked the capacity to explain their decisions \cite{elsherief-etal-2021-latent}. The advent of LLMs has revolutionized this domain by introducing models with deep semantic understanding and reasoning capabilities\cite{ghorbanpour-etal-2025-prompting}. However, this technological leap is characterized by a profound paradox: the widespread deployment of LLMs has simultaneously lowered the barrier to entry for malicious actors while raising the ceiling for defensive capabilities \cite{BARMAN2024100545}. Furthermore, the threat landscape itself has evolved in response to generative AI. We are no longer detecting only human-generated abuse; we are now tasked with identifying machine-generated disinformation, Fake Reviews, and attacks on social systems using LLM agents \cite{2025-legalsim-acl} \cite{10.1145/3726302.3730027} \cite{pmlr-v299-scama25a} \cite{meng2025largelanguagemodelshidden}. The dual-use nature of LLMs as both the sword for attackers and the shield for defenders is a recurring theme in the literature.

\subsection{Direct Zero-Shot and Few-Shot Classification}
The most immediate application of LLMs in abuse detection is the LLM-as-a-Judge paradigm, where general-purpose models are prompted to evaluate content directly \cite{gu2024surveyllmasajudge}. This approach bypasses the need for training task-specific models, relying instead on the model's pre-trained understanding of language and social norms \cite{NEURIPS2020_1457c0d6}. Research in this area has focused on benchmarking these capabilities against human baselines and identifying the specific failure modes of zero-shot reasoning in adversarial environments.

\subsubsection{The Efficacy of Foundation Models as Moderators}
Early experiments sought to benchmark foundation models against traditional supervised baselines like fine-tuned BERT and human annotation. The consensus emerging from this body of work is that large-scale proprietary models can achieve high agreement rates with human moderators on tasks involving explicit toxicity.

Zero-Shot Performance: In zero-shot settings, where the model is provided only with a prompt definition of the abuse category (e.g., ``Classify the following text as hate speech or not''), high-capacity models often outperform crowd-sourced workers in consistency, if not always in accuracy. For example, GPT-4 has been shown to achieve F1 scores exceeding 0.75 on standard benchmarks, rivaling the inter-annotator agreement of non-expert humans \cite{Kumar_AbuHashem_Durumeric_2024}.

Few-Shot Amplification: The introduction of few-shot examples i.e. providing 3 to 5 labeled instances of safe and unsafe content within the prompt context has been identified as a critical lever for performance. Research on Hate-Speech and Emotion Detection benchmarks indicates that few-shot prompting can elevate foundation models to state-of-the-art levels, effectively bridging the gap with fully fine-tuned specialist models \cite{bauer-etal-2024-offensiveness}. This capability is particularly valuable for long-tail abuse categories where training data is scarce, such as specific types of harassment in niche online communities.

\subsubsection{Advanced Prompt Engineering for Safety}
Researchers have developed sophisticated prompting strategies designed to guide the model's reasoning process and reduce sensitivity to artifacts\cite{badhe2026promptleveldistillationnonparametricalternative}.

Persona-Based Prompting: Instructing the model to adopt a specific persona such as an expert content moderator compliant with platform safety guidelines or a linguistic expert specializing in pragmatics has been shown to improve detection accuracy. This technique helps contextualize the model's analysis, particularly for categories like patronizing or condescending language which rely heavily on social power dynamics rather than explicit slurs \cite{kruschwitz-schmidhuber-2024-llm}.

Definition-Aware and Role-Playing Prompts: Rather than relying on the model's internal, opaque definition of hate speech, successful frameworks employ prompts that include explicit, granular definitions of the prohibited categories. For example, defining doxing precisely as the broadcasting of private or identifying information about an individual without their consent reduces ambiguity and improves consistency across different model architectures \cite{koh-etal-2024-llms}.

Chain-of-Thought (CoT) for Zero-Shot: While typically associated with complex reasoning, CoT is increasingly applied in zero-shot classification to force the model to justify its decision before outputting a label. This reasoning tracen acts as a buffer against surface-level keyword matching, allowing the model to distinguish between a slur used in a hateful context versus a reclaimed or educational context \cite{ghorbanpour-etal-2025-prompting}.

The trajectory of research suggests that while zero-shot classification is a powerful tool for rapid prototyping and data labeling, it lacks the robustness required for automated, high-stakes decision-making. The inherent safety filters of general-purpose models often conflict with the objective neutrality required of a classifier, leading to the over-refusal paradox. This limitation has driven the field toward the development of specialized, fine-tuned models that decouple safety capabilities from general generative abilities.

\subsection{Fine-Tuned Specialist Models}
While zero-shot models offer versatility, the practical constraints of deployment cost, latency, and strict adherence to policy have driven the development of Fine-Tuned Specialist Models. These are LLMs that have been instruction-tuned specifically for the task of content moderation, bridging the gap between the reasoning of large models and the efficiency of classifiers.

\subsubsection{The Llama Guard Ecosystem}
The most significant contribution to this category is the Llama Guard family of models, released by Meta, which shifts from implicit toxicity detection to taxonomy-based safeguarding [15]. Llama Guard is a Llama-based model fine-tuned on mixed clean and adversarial example and performs Input-Output Safeguarding to evaluate both the user's prompt for jailbreak prevention and the AI's response (for toxic content). A key innovation is its Zero-Shot Policy Adaptation feature, allowing administrators to pass a specific safety policy (e.g., ``Ban all mention of cryptocurrencies'') in the prompt without retraining, enabling flexibility for diverse use case \cite{inan2023llamaguardllmbasedinputoutput}. The family has evolved with the release of Llama Guard 3 Vision to detect multimodal abuse, such as hateful memes or embedded text, by utilizing visual instruction tuning to reason about the interplay between image and text \cite{chi2024llamaguard3vision}. Furthermore, Llama Guard 3 incorporates Tool-Use Security to protect against attacks designed to trick a code interpreter into executing malicious script \cite{chi2024llamaguard3vision}.

\subsubsection{Domain-Specific Challenges}
General-purpose safety models often struggle with the specific dynamics of user-AI interactions. The ToxicChat benchmark was developed to address this gap \cite{lin2023toxicchat}. Analyzing real-world user queries reveals that abuse in chatbot interactions differs significantly from social media toxicity. It often involves jailbreaking subtle, linguistically complex attempts to manipulate the model's rules rather than overt slurs. The process frequently requires overcoming subtle, complex linguistic efforts to manipulate the model's guidelines, rather than addressing obvious offensive language. ToxicChat demonstrates that models fine-tuned on standard social media datasets like the Jigsaw toxicity dataset perform poorly on these interactional attacks. Fine-tuning specifically on the ToxicChat dataset, which contains high-quality, human-annotated examples of jailbreaks and subtle toxicity, significantly improves detection performance \cite{lin2023toxicchat}. This underscores the principle that data quality dominates model scale in specialist applications; a smaller model fine-tuned on domain-specific data often outperforms a larger generalist model.

Beyond creating external guard models, researchers have explored Attribute Controlled Fine-tuning to inherently align the generation model itself \cite{meng-etal-2024-attribute}. This method involves fine-tuning the LLM on a corpus while applying a sequence-level constraint (e.g., minimizing toxicity scores). Unlike standard RLHF, which often leads to refusal behaviors, ACFT aims to teach the model to effectively handle toxic prompts by generating safe, non-toxic responses rather than simply shutting down \cite{meng-etal-2024-attribute}. However, the fine-tuning process itself is a vector for attack. Instruction Fine-Tuning Attacks involve injecting poisoned examples into the training data \cite{li2025detectinginstructionfinetuningattacks}. A subtle modification to just 1\% of the fine-tuning dataset can create backdoors specific trigger phrases that cause the model to output harmful content. Detecting these poisons is computationally expensive, often requiring influence function analysis to trace the impact of specific training examples on model behavior \cite{li2025detectinginstructionfinetuningattacks}. This reveals a critical vulnerability in the supply chain of fine-tuned models: if the dataset is crowdsourced or unverified, the safety alignment can be compromised at the source.

\subsection{Embedding-Based Detection}
While fine-tuned generative models offer interpretability, Embedding-Based Detection focuses on the latent vector space of language. By converting text into dense vector representations, researchers can leverage geometric properties to detect subtle forms of abuse that lack explicit keywords. This domain has been revolutionized by the application of Contrastive Learning (CL), which addresses the nuances of implicit hate speech.

\subsubsection{Latent Hatred and Contrastive Learning}
Explicit hate speech is easily detected via keyword matching, but implicit hate characterized by sarcasm, stereotypes, and coded language remains a significant challenge. The Latent Hatred benchmark highlights the failure of traditional models to capture these nuances  \cite{elsherief-etal-2021-latent}. To address implicit hate, researchers have applied Contrastive Learning to LLM embeddings, reshaping the vector space so implicit hate is topologically closer to explicit hate and further from benign text \cite{10.1145/3746275.3762209}. Frameworks like FiADD (Focused Inferential Adaptive Density Discrimination) augment the latent space by pulling implicit hate closer to its explicit counterparts, essentially teaching the model that subtle hate shares the same underlying intent as a slur \cite{Masud_2024}. Conversely, Causality-Guided Contrastive Learning (CCL) introduces causal intervention to disentangle hate intent from spurious correlations like specific dialect markers, thus improving generalization to new, unseen hate forms \cite{jiang-2025-learn}. Furthermore, Fusion Strategies explore integrating multiple information sources such as content, emotion, and context into the embedding space; integrating emotion helps the model distinguish between hateful anger and righteous indignation \cite{10.1145/3746275.3762209}. Crucially, studies show that fine-tuning just the embedding layers of general-purpose models like E5 or RoBERTa using these contrastive techniques can achieve State-of-the-Art (SOTA) performance on implicit hate detection, often surpassing much larger generative models in efficiency and accuracy \cite{elsherief-etal-2021-latent}.

\subsubsection{Dense Retrieval and RAG for Moderation}
The integration of Retrieval-Augmented Generation (RAG) has introduced Dense Retrieval strategies to abuse detection, allowing the system to use external database context instead of relying solely on the model's frozen weights. A pioneering framework is MetaTox, which builds a meta-toxic knowledge graph by having LLMs extract structured triplets (e.g., $<$Term X, implies, Derogatory Stereotype Y$>$) from toxic content; this retrieved toxic knowledge is then used as context for the LLM when classifying new posts \cite{zhao-etal-2025-enhancing-llm}. This method dramatically reduces false negatives in out-of-domain situations; for example, a new slang term can be immediately effective through retrieval without costly retraining. Furthermore, other research aims to capture uncertainty in dense retrieval by representing embeddings as probabilistic distributions rather than deterministic points, thereby better handling ambiguous queries and improving generalization to out-of-distribution abuse \cite{10.1145/3637870}.

The use of dense embeddings for user content raises significant privacy concerns. Inference-time context leakage is a documented risk where the embeddings used for retrieval may inadvertently encode sensitive user attributes that are not relevant to the moderation task\cite{mireshghallah2025positionprivacyjustmemorization}. This data leakage allows potential adversaries to reconstruct private information from the vector database. Consequently, the field is moving toward privacy-aligned representation learning, which seeks to minimize the mutual information between the embedding and sensitive attributes while maximizing the information relevant to abuse detection \cite{mireshghallah2025positionprivacyjustmemorization}.

\subsection{Network \& Narrative Analysis}
Abuse detection has historically been content-centric, analyzing individual posts in isolation. However, sophisticated actors employ Coordinated Inauthentic Behavior (CIB), creating networks of accounts to amplify disinformation, harass targets, or manipulate trends. Detecting this requires analyzing the structure of interactions, not just the text. The integration of LLMs with Graph Neural Networks (GNNs) represents a Neuro-Symbolic leap in this field.

\subsubsection{LLM-Enhanced Graph Neural Networks}
Traditional GNNs rely on structural features (e.g., ``who follows whom'') or simple text features (e.g., Bag-of-Words). However, these conventional features frequently fail to capture the subtle linguistic indicators of coordination. Consequently, new architectures have emerged to fuse the semantic depth of LLMs with the relational power of GNNs. FraudSquad is a notable framework developed for detecting LLM-generated spam reviews and coordinated attacks\cite{liu2025detectingllmgeneratedspamreviews}. This approach enhances node representations within a review graph by integrating dense text embeddings derived from a pre-trained LLM. The augmented nodes are subsequently processed by a Gated Graph Transformer, which is trained to discern anomalies in the patterns of posting. For example, it can identify a set of accounts that exhibit not only temporal posting synchronicity (structural feature) but also leverage analogous LLM-generated linguistic structures (semantic feature). By employing this dual-view mechanism, FraudSquad has achieved a 44\% enhancement in precision compared to state-of-the-art baselines.

Social-LLM and MLED address the inherent difficulty of scaling graph analysis to networks comprising billions of edges\cite{sci7040138}\cite{10.1145/3746027.3755245}. Social-LLM mitigates this challenge through an edge-based sampling technique, training on localized subgraphs to learn user representations that effectively synthesize content cues (such as bios and posts) with network homophily. Concurrently, MLED (Multi-level LLM Enhanced Graph Fraud Detection) employs LLMs to extract external knowledge from text specifically, by reasoning about the nature of relationships and subsequently embeds this semantic knowledge into the graph edges. This mechanism significantly enhances the model's capacity to distinguish genuine communities from artificially created ones.

\subsubsection{Narrative Clustering and Disinformation}
LLMs are exceptionally skilled at summarization and thematic extraction, making them ideal for identifying Disinformation Narratives. Rather than fact-checking every individual post, researchers use LLMs to extract narrative embeddings from thousands of posts to identify common themes\cite{info16040297}. Previously, authors have demonstrated the use of Knowledge Graphs (KGs) to map these narratives\cite{mandravickaite-2025-narrative}. By analyzing the structure of the KG, researchers can distinguish between disinformation which often exhibits fragmented, abstract connections and emotional language and trustworthy news which shows balanced, concrete relationship patterns. A major challenge is tracking these narratives as they jump across platforms. Coordinated Sharing Behavior (CSB) detection algorithms analyze the synchronization of narrative sharing. If a cluster of accounts on X and TikTok share the same complex narrative simultaneously, it is a strong signal of inauthentic coordination\cite{10.1145/3696410.3714698}.

\subsubsection{Stylometric Analysis and Bot Detection}
The rise of LLM-generated content has rendered traditional bot detection based on grammar errors or simple repetition obsolete. This has necessitated the evolution of Stylometric Analysis. StyloCPA uses LLMs to analyze the writing style of a user's timeline over time. It detects change points such as sudden shifts in entropy, perplexity, or syntactic structure that indicate an account has been sold or automated\cite{kumarage2023stylometricdetectionaigeneratedtext}. This forensic approach is critical for detecting cyborg accounts that alternate between human and AI operation. Researchers are enhancing GNNs with LLM-generated node features. Traditional GNNs rely on network topology (who follows whom). New frameworks like BotLGT and LGB use LLMs to generate rich semantic embeddings for each user node\cite{11015729} \cite{10.1609/aaai.v39i13.33575}. The GNN then aggregates these semantic features along the graph edges to detect sybil clusters of fake accounts that are both structurally and semantically correlated. BotHash is a training-free approach that leverages simplified user representations for approximate nearest-neighbor search, effectively differentiating between human and bot accounts even when the bots use state-of-the-art LLMs to generate content\cite{dipaolo2025bothashefficienttrainingfreebot}.

\subsection{Complex Reasoning and Agents}
Traditional classifiers act as black boxes, they assign a label but cannot explain it. Reasoning-aware systems leverage the generative nature of LLMs to articulate the rationale behind a decision, enabling the detection of implicit abuse and supporting human moderators.

\subsubsection{Chain-of-Thought (CoT) for Explanation}
Implicit hate speech often relies on metaphors, irony, or world knowledge e.g., ``He's one of those people''. Standard classifiers often miss this. Chain-of-Thought (CoT) prompting forces the model to generate an intermediate reasoning step before the final label. For example, the model might output ``Reasoning: The phrase 'one of those people' in this context refers to a specific minority group and implies they are undesirable. Verdict: Hate Speech.'' This explicit reasoning trace significantly improves performance on nuanced datasets compared to direct classification\cite{ghorbanpour-etal-2025-prompting}. The HateCOT dataset serves as a benchmark for this capability. It contains over 52,000 samples of abuse annotated not just with labels, but with detailed CoT explanations. Pre-training models on this explanation-rich data has been shown to improve their ability to generalize to new, unseen domains of abuse, effectively teaching the model how to think about toxicity rather than just memorizing patterns\cite{nghiem-daume-iii-2024-hatecot}.

\subsubsection{Agentic Moderation}
Moving beyond single-model prompting, Agentic Moderation involves systems of multiple LLM agents working in concert. Research demonstrates that a multi-agent debate system where one agent acts as a prosecutor and another as a defender, with a third judge agent making the final call reduces hallucinations and improves decision quality on ambiguous content\cite{eaa25}. This mimics a human jury process. Qiyuan Zhang et al. propose introducing crowd responses to compare with candidate responses\cite{zhang-etal-2025-crowd}. This comparative approach exposes deeper details and guides the LLM-as-a-Judge to provide more comprehensive judgments than it would in isolation. Agentic systems can also be equipped with external tools. AutoRedTeamer describes agents that can query external databases or perform web searches to verify the context of a potential disinformation claim, effectively performing a specialized investigation for each flagged item\cite{zhou2025autoredteamerautonomousredteaming}.

\section{Stage III: Review and Appeals}

While the detection stage of the Abuse Detection Lifecycle (ADL) prioritizes throughput and recall, the Review and Appeals stage necessitates a strategic shift toward precision, fairness, and interpretability. Human moderators operating at this stage frequently face high cognitive loads induced by the volume of flagged content and the requirement to interpret complex, evolving safety policies \cite{huang2025content}. Consequently, the integration of LLMs moves beyond simple classification to function as Cognitive Support Assistants. In this capacity, LLMs are tasked with bridging the gap between automated detection and human adjudication by generating policy-grounded justifications, summarizing evidence, and facilitating the appeals process to enhance the overall legitimacy of the governance system.

\subsection{Explanation Generation and Reasoning}
A critical limitation of traditional black-box classifiers is the reasoning gap between a predicted label and the specific policy violation. Standard fine-tuning on human annotations often fails to capture the implicit nuances of abuse because human labels typically lack the intermediate reasoning required to explain why a specific phrase is abusive \cite{yang2023hare}. The HARE (Explainable Hate Speech Detection with Step-by-Step Reasoning) framework demonstrates that LLMs can bridge this gap by generating rationales that explicitly identify the target group and implied meaning of a post before assigning a classification \cite{yang2023hare}. This sequential generation of rationales significantly enhances the interpretability of the decision for human reviewers compared to standard classification outputs.

To ground these explanations in factual contexts, models often require access to external knowledge. Research on GLlama Alarm indicates that the implicit knowledge within off-the-shelf LLMs is frequently insufficient for multi-class abuse detection, particularly when distinguishing between levels of offensiveness \cite{di2025detection}. By integrating external knowledge bases spanning encyclopedic, commonsense, and temporal-linguistic data, systems can generate structured explanations that are semantically and syntactically aligned with human reasoning \cite{di2025detection}. This Knowledge-Guided approach mitigates lexical overfitting, where models flag terms without context, and aids in identifying nuanced abuse by referencing specific real-world events or slurs \cite{di2025detection}.

To further structure this reasoning, recent architectures propose embedding LLMs within standardized analytical processes, such as the Question-Option-Criteria (QOC) framework \cite{jansen2025increasing}. Rather than treating the LLM as an opaque reasoning engine, this architecture separates the unexplainable reasoning space of the model's internal weights from an explainable process space \cite{jansen2025increasing}. By constraining the LLM to populate a transparent decision matrix evaluating options against explicit criteria, the system produces an auditable decision trace \cite{jansen2025increasing}. This ensures that the generated justification follows a consistent, logical structure rather than a stream-of-consciousness narrative, thereby enhancing accountability.

\subsection{Augmenting Human Moderators}
The primary objective of generating explanations in the ADL is to assist human moderators in making faster, more accurate, and more consistent judgments. The integration of LLMs into review workflows supports a paradigm shift from optimizing solely for accuracy to optimizing for legitimacy \cite{huang2025content}. Under this framework, moderation cases are categorized into easy and hard cases. For easy cases, LLMs prioritize speed and automation. However, for hard cases involving complex values, cultural context, or ambiguous intent, the LLM’s role is to provide high-quality reason-giving to assist human experts \cite{huang2025content}. This distinction ensures that human cognitive effort is reserved for cases where subjective judgment and nuanced policy interpretation are strictly necessary.

Practical implementations, such as the LLM-Mod framework, utilize LLMs to interpret community guidelines directly \cite{kolla2024llm}. By prompting the model with specific platform policies, the system can identify rule violations that go beyond keyword matching, such as detecting speculative answers in historical forums \cite{kolla2024llm}. While these models may struggle with consistency in final verdicts, they excel at defining key terms and identifying problematic segments of text \cite{kolla2024llm}. This capability allows LLMs to serve as effective triage assistants that highlight relevant policy clauses for human review, reducing the cognitive burden of mapping content to complex rule sets \cite{kolla2024llm}.

LLMs also play a role in monitoring the quality of the reviews themselves. Analogous to systems deployed in academic peer review, Review Feedback Agents can analyze the text of a human moderator’s decision to detect vague critiques, misunderstandings of content, or unprofessional tone \cite{thakkar2025can}. In large-scale randomized control trials, such agents have been shown to increase the specificity and actionability of reviews by prompting reviewers to clarify their reasoning \cite{thakkar2025can}. Implementing similar meta-review agents in trust and safety operations can ensure that the justifications provided to users during the appeals process are high-quality, specific, and actionable.

\subsection{Facilitating Appeals and User Agency}
The final component of the Review stage is the interface with the user. Legitimacy in platform governance requires procedural justice, granting users the ability to understand and contest decisions \cite{huang2025content}. LLMs facilitate this by translating complex policy violations into natural language explanations personalized for the user. Moreover, conversational agents can manage the initial stages of the appeal process, allowing users to provide context or rebuttal in a dialogic format \cite{huang2025content}. By processing these user inputs and summarizing them for human reviewers, LLMs reduce the friction of the appeals process, ensuring that hard cases receive the necessary deliberation while maintaining the scalability of the safety system \cite{huang2025content}.

\section{Stage IV: Auditing and System Governance (The Consistency Challenge)}

Auditing constitutes the final layer of the Abuse Detection Lifecycle (ADL) and plays a crucial role in ensuring that a moderation system operates fairly, consistently, and in alignment with organizational governance expectations. While labeling, detection, and review determine how individual judgments are produced, auditing offers a system-level evaluation of whether these judgments are legitimate, non-discriminatory, stable over time, and resilient to adversarial manipulation. In the context of LLM-enabled moderation, where models are updated frequently and behave non-deterministically, auditing transitions from a periodic assessment to an integrated, continuous governance practice.

This section surveys the literature across six domains central to auditing LLM-driven abuse detection systems: (1) fairness and representational harm audits, (2) longitudinal and drift monitoring, (3) explanation and legitimacy audits, (4) adversarial and jailbreak stress audits, (5) governance and regulatory auditing, and (6) multimodal auditing. Together, these domains form the foundation of modern auditing infrastructures for LLM-based moderation pipelines.

\subsection{Fairness, Bias, and Representational Harm Audits}
Fairness audits evaluate whether moderation systems exhibit disparate behavior across demographic or identity groups. Dutta (2024) analyzes toxicity elicitation across 1,266 identity groups, uncovering systematic harms toward marginalized populations by stress-testing the model's safety filter \cite{ijcai2024p801}. Their findings demonstrate that LLMs can unintentionally reintroduce discriminatory patterns, making identity-aware evaluation a central component of auditing. Zangl (2025) similarly observe that AI-driven toxicity detection tools deployed in civic platforms inconsistently handle content involving minority groups and often lack clear transparency or explanation mechanisms, exacerbating inequities \cite{Zangl2025AMA}. A Comprehensive Review of LLM-based Content Moderation warns of an emerging algorithmic monoculture, in which a dominant LLM worldview shapes moderation norms globally. Together, these results underscore the necessity of fairness-focused audits that examine how LLM-based moderation behaves across identities, contexts, and cultural settings.

\subsection{Longitudinal and Drift Auditing}
Longitudinal audits examine whether moderation systems behave consistently over time. Pozzobon (2023) demonstrates substantial temporal instability in toxicity predictions, showing that Perspective API scores vary significantly across evaluation periods and alter model rankings \cite{pozzobon-etal-2023-challenges}. Chen (2025) complements this by noting that moderation models face the risk of catastrophic forgetting when adapting to shifting policy landscapes and evolving adversarial threats \cite{CHEN2025114689}. This evidence indicates that moderation systems require continuous monitoring of refusal patterns, toxicity thresholds, and category-level sensitivities to maintain consistency across time and model versions.

\subsection{Explainability, Transparency, and Legitimacy Audits}
Explanation-oriented auditing evaluates whether moderation decisions are justified in ways that uphold procedural fairness. Huang (2025) argues that legitimacy not accuracy is central to evaluating moderation outcomes, especially for hard cases, which require reason-giving and transparency \cite{huang2025content}. Zangl (2025) highlighted related issues in toxicity detection systems, noting inconsistent or incomplete toxic-span explanations and the absence of standardized interpretability measures \cite{Zangl2025AMA}. Gorwa (2020) further emphasized long-standing transparency deficits in algorithmic moderation pipelines, which undermine accountability and trust \cite{doi:10.1177/2053951719897945}. These works collectively motivate auditing practices that evaluate rationale fidelity, policy alignment, and whether explanations meet governance requirements.

\subsection{Adversarial, Jailbreak, and Guardrail Stress Audits}
Adversarial audits assess resilience to attempts to circumvent or exploit moderation boundaries. Dutta(2024) iterative toxic elicitation tests reveal how subtle prompt variations can bypass LLM guardrails and induce harmful outputs \cite{ijcai2024p801}. Irtiza (2024) complements this by conducting a multilingual adversarial audit that exposes persistent jailbreak vulnerabilities across models including ChatGPT, Gemini, and Mistral \cite{10835584}. In response, it proposes a defense framework utilizing sliding-window analysis and RAG-based evaluation to detect these harmful intents. These findings demonstrate that adversarial evaluation must be continuous and integrated directly into governance practices.

\subsection{Governance, Regulatory, and Sociotechnical Audits}
Algorithmic Content Moderation: Technical and Political Challenges assess whether moderation aligns with legal, ethical, and platform-level governance principles. Gorwa (2020) identify structural opacity, fairness challenges, and accountability gaps in algorithmic content moderation systems, many of which are amplified by LLMs \cite{doi:10.1177/2053951719897945}. Chen (2025) review situates these concerns within the broader context of LLM deployment, identifying risks related to ethical consistency, monoculture effects, and insufficient procedural oversight \cite{CHEN2025114689}. Huang (2025) further argues that governance-compliant moderation must emphasize transparency, reason-giving, and user involvement throughout the process \cite{huang2025content}. These works together establish the foundation for governance-centered audits that go beyond accuracy to ensure procedural legitimacy and equitable enforcement.

\subsection{Multimodal Auditing}
With the rise of image, video and mixed-media abuse, multimodal auditing has become essential. Zangl (2025) shows that multimodal toxicity detection tools deployed in civic platforms often lack explainability and struggle with context fusion \cite{Zangl2025AMA}. Chen (2025) similarly identifies multimodal evaluation as an emerging challenge, highlighting the need for benchmarks that capture semantic alignment, cross-modal biases, and vulnerability in multimodal fusion systems \cite{CHEN2025114689}. These findings point to the need for robust multimodal audit frameworks that evaluate cross-modal consistency and ensure that harmful non-text content is handled with the same rigor as textual abuse.

\section{Challenges and Future Research Directions}

The transition of LLMs from experimental benchmarks to operational safety pipelines introduces a distinct set of engineering and governance challenges that are often obscured by high performance on static leaderboards. While LLMs demonstrate unprecedented semantic reasoning, their integration into real-world abuse detection systems is constrained by the production gap: the fundamental tension between the computational intensity of massive models and the strict latency, throughput, and cost-efficiency requirements of platform-scale moderation. Furthermore, deploying these models introduces critical risks regarding reliability, including the stochastic nature of outputs, susceptibility to adversarial 'jailbreaks', and the potential for unfaithful explanations that diverge from the model's internal reasoning. This section analyzes these systemic barriers and outlines a future research agenda aimed at operationalizing LLMs as efficient, robust, and accountable components of digital safety infrastructure. To provide a granular analysis, we map these challenges across the four stages of the Abuse Detection Lifecycle (ADL): Labeling, Detection, Review, and Auditing.

\subsection{Stage I: Labeling}
\textbf{Challenges: Circular Bias, Subjectivity, and Prompt Sensitivity.} A primary challenge in deploying LLMs as data amplifiers is the risk of encoding systemic biases into the labeled data. Research indicates that LLM-generated labels often reflect the specific alignment strategies of the model rather than objective ground truth. For instance, instruction-tuned models (e.g. Flan-T5) tend to under-predict abusive labels due to training on negative-majority corpora, whereas RLHF-aligned models (e.g., LLaMA-2) frequently over-predict abuse, err on the side of caution, and exhibit over-refusal behaviors \cite{zhang2025llm}. This introduces a circular bias where synthetic datasets generated by these models such as those for cyberbullying, may inadvertently amplify the model’s latent safety constraints or refusals rather than accurately reflecting real-world toxicity \cite{202407.0411}. Furthermore, static text labeling often fails to capture the intrinsic ambiguity of abusive communication. Abuse is highly context-dependent; text-only annotations miss critical paralinguistic cues like tone and pitch, which are essential for distinguishing between neutral statements and verbal abuse in short conversations \cite{10976181}. Additionally, recent evaluations reveal that LLMs are highly sensitive to prompt artifacts; a model may classify the same implicitly abusive sentence as harmful or neutral depending entirely on the prompt template used, undermining the reliability of zero-shot labeling \cite{jaremko-etal-2025-revisiting}.

\textbf{Future Research Directions: Making Systems Smarter, Personal, and Human-Guided.} To solve these problems, future research needs to move toward systems that are flexible and keep humans in the loop. While automated tools can generate complex data (like Q\&A pairs), they are still prone to hallucinating or making up facts. This means we cannot automate everything; we need human experts to verify the AI's output to guarantee the data is high quality \cite{rocha2025data}. Frameworks like LLM-Sentry prove that it is much better to maintain a dynamic, human-curated list of threats that evolves over time, rather than relying on a frozen model that cannot adapt to new attacks \cite{10835584}.

There is also a critical need for personalized labeling. We know that different people perceive abuse differently. Studies show that when we teach the model about a user’s specific personality traits (like how sensitive or rational they are), the model becomes much better at detecting harm relevant to that person. However, we have to be careful: future research needs to make sure these systems don’t become ``over-personalized'' or inconsistent, where the same rule is applied unfairly to different people \cite{yao-etal-2024-personalised}.

Finally, research should focus on saving money and combining models. One promising method is Aggregation-of-Subjectivity, which just means asking several different LLMs to vote on a label. The majority vote from these models is often as accurate as a human expert \cite{horych-etal-2025-promises}. To save on costs, we can use Retrieval-Augmented Generation (RAG). Instead of feeding the model thousands of examples, RAG picks only the few most relevant examples for the specific post being checked. This makes it cheap enough to monitor huge streams of data continuously \cite{10.1145/3604237.3626891}. Lastly, instead of just saying safe or unsafe, future systems should use semantic augmentation using LLMs to explain why something is harmful or to list specific trigger words. This extra context helps smaller, faster models understand the nuance of abuse much better \cite{Meguellati_Zeghina_Sadiq_Demartini_2025}.

\subsection{Stage II: Detection}
\textbf{Challenges: Production Gap, Safety Paradox, and Over-Refusal.}
The transition from static classifiers to LLMs in abuse detection has introduced a complex set of operational and robustness challenges. While LLMs offer superior semantic understanding, their deployment is hindered by significant Production Gaps and vulnerability to sophisticated adversarial attacks. 

 A primary barrier to operationalizing LLMs is the tension between reasoning depth and computational cost. Unlike traditional BERT-based classifiers that process inputs in parallel, LLMs rely on sequential autoregressive decoding, which creates substantial latency bottlenecks unfit for high-velocity real-time moderation. Research presented at ACM ICAIF '23 quantifies quantifies this economic disparity, demonstrating that while frontier models like GPT-4 achieve high accuracy in few-shot scenarios, the cost per inference is orders of magnitude higher than distilled baselines\cite{10.1145/3604237.3626891}. This Production Gap forces platforms into a difficult trade-off: utilizing capable but expensive models restricts throughput, while cheaper models often fail to grasp the nuances of complex abuse. Recent SafeRoute framework further identifies that static deployment of large safety guards incurs unnecessary overhead, as a significant portion of user traffic is easy and does not require the reasoning depth of a multi-billion parameter model\cite{lee-etal-2025-saferoute}. 

The democratization of generative AI has fundamentally altered the threat landscape, leading to a Safety-Paradox where better defenses spur more sophisticated attacks. The ToxicChat benchmark highlights that real-world user-AI interactions are replete with jailbreaks which are complex, role-playing prompts designed to bypass safety filters that standard toxicity classifiers fail to detect\cite{lin2023toxicchat}. Unlike explicit hate speech, these adversarial inputs exploit the model's instruction-following nature, effectively turning its own capabilities against it. Conventional monitoring systems, trained on static social media comments, struggle to generalize to these dynamic, multi-turn adversarial strategies, leaving platforms vulnerable to stealth abuse that does not trigger keyword filters.

Despite the promising benchmarks, the deployment of general-purpose LLMs as safety classifiers is hindered by significant reliability issues. A primary concern documented in recent literature is the phenomenon of Over-Refusal. Commercial foundation models are heavily aligned via Reinforcement Learning from Human Feedback, or RLHF to avoid generating or engaging with harmful content. When repurposed as classifiers, this safety alignment often backfires. The models become overly cautious, flagging benign but controversial content such as discussions about historical violence, medical terminology, or fictional weapons as unsafe. Studies evaluating Llama-2 and GPT-4 found a persistent tendency to prioritize refusal over nuanced analysis, leading to high False Positive Rates (FPR) on safe prompts that merely resemble unsafe ones \cite{zhang-etal-2024-dont-go}. LLMs exhibit a troubling sensitivity to input artifacts subtle variations in the prompt that should not affect the semantic classification but do. For instance, the presence of apologetic language (e.g., ``I'm sorry, but...'') in a response can skew an LLM judge's safety evaluation, leading it to rate the content as safer than it is \cite{chen-goldfarb-tarrant-2025-safer}. Furthermore, the specific choice of few-shot examples can introduce biases; if the few-shot examples all share a specific syntactic structure, the model may overfit to that structure rather than the semantic meaning of abuse \cite{bauer-etal-2024-offensiveness}.

\textbf{Future Research Directions:  Red Teaming, Neuro-Symbolic and Graph-Enhanced Detection, and Efficient and Specialized Guardrails.}
To address these systemic limitations, the next generation of detection architectures must
move beyond static classification toward adaptive, context-aware, and agentic systems.

Future research must prioritize continuous adaptation over static benchmarking. The
AutoRedTeamer framework introduces the concept of lifelong red teaming,
where autonomous multi-agent systems continuously generate novel attack vectors to probe
target models\cite{zhou2025autoredteamerautonomousredteaming}. By integrating these automated adversaries directly into the training loop, defense systems can be immunized against evolving threats in real-time. This shifts the
paradigm from reacting to past abuse to proactively discovering and mitigating vulnerabilities such as new jailbreak templates or visual prompt injections before they can be exploited at scale.

Detecting sophisticated abuse, such as coordinated fraud rings or disinformation campaigns, requires analyzing the structural relationships between actors, not just isolated text. Emerging research advocates for Neuro-Symbolic architectures that fuse the semantic reasoning of LLMs with the structural insights of Graph Neural Networks (GNNs). For instance, the MLED framework (ACM Multimedia 2025) demonstrates that using LLMs to extract semantic rationales and embedding them into a GNN significantly improves the detection of fraudsters who camouflage themselves within benign communities\cite{10.1145/3746027.3755245}. Similarly, Social-LLM shows that modeling network homophily alongside user content provides a robust signal for identifying coordinated inauthentic behavior that text-only models miss\cite{sci7040138}.

To bridge the production gap, research should focus on distilling the reasoning capabilities of large foundation models into smaller, specialized guardrails. Architectures like Llama Guard illustrate the effectiveness of fine-tuning compact models specifically for safety taxonomies rather than general capabilities\cite{10.5555/3737916.3738473}. Future directions include leveraging Retrieval-Augmented Generation (RAG) to allow these smaller models to dynamically reference updated policy guidelines. This Zero-Shot Policy Adaptation enables detection systems to instantly adapt to new forms of abuse such as novel slurs or emerging hate symbols by simply updating the retrieval index, avoiding the expensive and slow process of full model retraining.

\subsection{Stage III: Review And Appeals}
\textbf{Challenges: Plausibility Discrepancies, Persuasion Bias, and Sensitivity Imbalances.} The integration of LLMs into review workflows faces a critical obstacle in the divergence between plausibility and faithfulness where models generate convincing rationales that are disconnected from their actual decision making processes. \cite{turpin2023language} demonstrates that Chain-of-Thought reasoning can be systematically unfaithful by rationalizing incorrect predictions to align with biases present in the prompt. This discrepancy creates a scenario where explanations appear logically sound to human moderators despite being factually ungrounded. This risk is amplified by persuasion bias in human and AI collaboration. \cite{wang2023evaluating} highlights that the high fluency of LLM generated text often misleads moderators into rating incorrect explanations as logical which significantly increases the error acceptance rate. Furthermore, operationalizing these models requires navigating the trade off between under flagging and over blocking. \cite{zhang2025llm} identifies that instruction tuned models frequently under-predict abuse while models fine-tuned with human feedback tend toward excessive sensitivity and hallucinate violations.

\textbf{Future Research Directions: Honest Explanations, Cognitive Forcing, and Dynamic Evaluation.} Addressing these risks requires shifting focus from surface level fluency to honest explanation mechanisms that are causally linked to internal model states. \cite{di2025detection} and \cite{jansen2025increasing} suggest structuring reasoning processes to ensure explanations are grounded in external knowledge or transparent criteria which allows operators to verify if a policy violation explanation truly reflects the reasoning used to flag the content. To mitigate persuasion bias, \cite{buccinca2021trust} proposes interventions such as cognitive forcing functions or side by side counter explanations that encourage critical thinking rather than passive acceptance of machine outputs. Finally, \cite{tu2025ode} indicates the need for protocols like Open-Set Dynamic Evaluation to continuously test against evolving abuse concepts and prevent the gaming of static benchmarks.

\subsection{Stage IV: Auditing}
\textbf{Challenges: Hidden Models, Broken Trust, and Political Governance.} A major problem with testing these AI models is that they are often black boxes, we cannot see how they work on the inside. Companies frequently change the rules or update their models without telling anyone. This means a safety test that worked yesterday might fail today, making it impossible for researchers to compare results over time or trust the findings \cite{pozzobon-etal-2023-challenges}. But the problem isn't just technical; it is also political. Being accurate isn't the same as being right. We spend too much time looking at math scores (like F1 metrics) and not enough time asking if the decisions are socially fair. As argued in \cite{doi:10.1177/2053951719897945, badhe2026longtailknowledgelargelanguage}, automated systems often lack the cultural context to make fair decisions, which creates a gap between what the computer thinks is safe and what users actually accept. If users feel the AI is hiding legitimate opinions, the system loses its authority, a conflict explored in \cite{huang2025content}. This lack of transparency is especially dangerous for civic platforms where open debate is necessary; without clear rules, toxicity detection can accidentally silence important political discussions \cite{Zangl2025AMA}.

Finally, the models themselves are inconsistent. Different models make different kinds of mistakes depending on how they were taught. Some models (trained with human feedback) are so scared of being offensive that they refuse to answer even harmless questions (over-refusal), while others might miss insults entirely. As shown in \cite{zhang2025llm}, these inconsistencies make it hard to grade them fairly. Even worse, simple changes in wording can confuse the model; it might catch an insult one time but miss the same hidden insult the next time just because the prompt changed slightly \cite{jaremko-etal-2025-revisiting}.

\textbf{Future Research Directions: Explanations, Stress Tests, and Security.} To fix these issues, future audits need to do more than just say safe or unsafe. The AI should be able to explain why it made a decision. Research shows that we need to move from simple detection to Chain-of-Thought explanations. When we teach the model to ``show its work'' and justify its labeling, it helps human reviewers understand the error and trust the system \cite{di-bonaventura-etal-2025-detection}. We also need to stop using easy tests. Humans are good at spotting obvious slurs, but AI struggles with hidden meaning. We should use difficult, computer-generated examples designed to trick the model. By using massive datasets specifically built to include adversarial (tricky) and implicit hate speech, we can find weaknesses that standard tests miss \cite{hartvigsen-etal-2022-toxigen}. This must include specific stress tests that look for racism, antisemitism, and misogyny to make sure the safety locks are actually working for vulnerable groups, rather than just checking for general bad words \cite{ijcai2024p801}.

Lastly, auditing needs to include security. We have to check if people can break the model with trick questions (jailbreaks). The best way to do this is to keep humans involved in the process to constantly update the security checks, rather than testing the model once and forgetting about it. As proposed in \cite{10835584}, this dynamic approach allows the audit to evolve as fast as the attackers do. This shift toward continuous, clear, and thorough checking is essential for the future of AI \cite{binh2025transforming}.

\section{Conclusion}
The arrival of LLMs marks the most significant change in online safety in over a decade. We are moving away from old systems that simply flagged bad words to intelligent agents that can reason, understand context, and analyze images alongside text. This paper explored the entire Abuse Detection Lifecycle (ADL) and found that LLMs transform every stage of the job. In the Labeling stage, they act as data amplifiers that solve the cold start problem by generating training data when humans cannot keep up. In detection, they function as specialized experts that handle the grey areas such as sarcasm or cultural nuance that smaller, faster models often miss. In the Review stage, they support overwhelmed human moderators by summarizing complex evidence and writing clear, policy-based explanations. Finally, in Auditing, they act as adversarial stress-testers that constantly attack our systems to find security holes before bad actors do.

However, this new power creates a Safety Paradox. The same tools that help defenders catch hate speech also lower the barrier for attackers to generate sophisticated abuse, scams, and disinformation at a massive scale. Furthermore, we found that LLMs are not neutral machines; they carry hidden biases depending on how they were trained. For example, some models are instruction-tuned and tend to miss abuse, while others trained with human feedback suffer from over-refusal, where they are so scared of being offensive that they block harmless content. This creates a risk of circular bias, where the model learns bad habits from its own generated data, reinforcing these errors over time. Additionally, these models are huge and expensive to run, creating a production gap where they are often too slow or costly to check every single post on a large platform.

To solve these problems, future research must shift its focus in three specific ways. First, we must move beyond simple accuracy scores and focus on legitimacy. It is not enough for the computer to be right; it must explain why it made a decision so that users feel treated fairly. Second, we need to build hybrid architectures to save money and time. Instead of using one giant model for everything, platforms should use small, fast models to scan most content, reserving the powerful LLMs only for the hardest cases. Techniques like Retrieval-Augmented Generation (RAG) can help here by allowing models to look up policy rules cheaply instead of memorizing them. Finally, security must be continuous. Because attackers are always finding new jailbreaks to trick the AI, our auditing systems cannot sleep. We need dynamic frameworks that keep humans in the loop to constantly update the system's defenses. Ultimately, AI is a powerful assistant, but it must remain part of a human-led system that prioritizes transparency, safety, and trust.

\section*{Acknowledgments}

All study design, literature review, synthesis, and writing were conducted by the authors. Generative AI tools (Gemini) were used only for grammar checking, and proofreading during the final polishing of the manuscript. No generative AI system was used to interpret prior work, or draw conclusions. The authors reviewed and approved all final text and remain fully responsible for the content of the paper.

\bibliographystyle{IEEEtran}
\bibliography{references}

\begin{IEEEbiographynophoto}
{SURAJ KATH} is a Software Engineer at Google, specializing in Machine Learning and abuse prevention within the Google Play Monetary Abuse Infrastructure team,. He holds an M.S. from the School of Computing at the University of Utah. His career has focused on applying machine learning to large-scale unstructured data, ranging from public health surveillance to digital security,. At Google, he develops advanced models to detect adversarial behaviors and protect the platform ecosystem. He is a Fellow of the British Computer Society (FBCS) and a Distinguished Fellow of the Soft Computing Research Society (SCRS)
\end{IEEEbiographynophoto}

\vskip 0pt plus -1fil

\begin{IEEEbiographynophoto}
{SANKET BADHE} is a seasoned Machine Learning Engineer with over 10 years of experience specializing in AI Security, large-scale ML systems, and LLM applications. He currently leads key ML initiatives at Google for Youtube shopping. Sanket holds a Master’s in Data Science from Rutgers University and a B.Tech from IIT Roorkee, with prior experience at Tinder, TikTok, Oracle etc. Sanket is a recognized expert in his field, Sanket’s research has been published in ACL, CAMLIS, and IEEE etc. Beyond his technical contributions, he is an active voice in the AI community, frequently delivering talks on LLM safety and industrial use cases at premier conferences such AI4 Conference, MLOps World etc.
\end{IEEEbiographynophoto}

\vskip 0pt plus -1fil

\begin{IEEEbiographynophoto}
{PREET SHAH} received the B.Tech. degree in Information Technology from Nirma University, India, in 2019, and the M.S. degree in Computer Science from Northeastern University, Boston, USA, in 2021. He is currently a Software Engineer with Google, Mountain View, USA, where he focuses on developing scalable infrastructure solutions for the Google Ads Anti-abuse team. His professional interests include distributed systems, large-scale data processing, and the application of Large Language Models (LLMs). Mr. Shah was a recipient of the Google Ads Privacy and Safety Engineering Excellence award in 2022 and the Google Perfy Award in 2024.
.
\end{IEEEbiographynophoto}

\vskip 0pt plus -1fil

\begin{IEEEbiographynophoto}
{ASHWIN SAMPATHKUMAR} is an Engineering Manager at Google, specializing in AI/ML and workflow automation within the Legal Engineering domain. He holds a B.Tech. in Information Technology from Anna University and a PGP in AI/ML from UT Austin. Throughout his 18-year career, including leadership tenures at Salesforce and Signet Jewelers, he has focused on building data-driven systems and predictive ML models at scale. He is a recipient of the Google Core Impact Tech Award in 2024 and the Motorola Scholar (FAER) Award in 2006. His expertise lies in agentic automation, event-driven SOA architecture, and building next generational solutions.
\end{IEEEbiographynophoto}

\vskip 0pt plus -1fil

\begin{IEEEbiographynophoto}
{SHIVANI GUPTA} is a Staff Software Engineer at Google. She has led teams across Google Ads and Cloud, building large-scale systems and ML models that impact multi-billion dollar ARR. She completed her Bachelor's degree at Delhi Technological University, India in 2016.
\end{IEEEbiographynophoto}

\EOD

\end{document}